%% file: autoconj.tex
\documentclass{article}
\input{preamble/preamble}

\input{preamble/preamble_acronyms}
\input{preamble/preamble_math}

\input{preamble/preamble_tikz}
\input{_minted-autoconj/default.pygstyle}
\input{_minted-autoconj/default-pyg-prefix.pygstyle}

 \usepackage[final]{neurips_2018}

\usepackage{ragged2e}

\usepackage[draft]{minted}

\title{Autoconj: Recognizing and Exploiting Conjugacy Without a Domain-Specific Language}

\author{
Matthew D. Hoffman\thanks{equal contribution} \\
  Google AI \\
  \texttt{mhoffman@google.com} \\
  \And
  Matthew J Johnson\textsuperscript{*} \\
  Google Brain \\
  \texttt{mattjj@google.com} \\
  \And
  Dustin Tran \\
  Google Brain \\
  \texttt{trandustin@google.com}
}

\begin{document}

\maketitle

\begin{abstract}
Deriving conditional and marginal distributions using conjugacy
  relationships can be time consuming and error prone.  In this paper,
  we propose a strategy for automating such derivations. Unlike
  previous systems which focus on relationships between pairs of
  random variables, our system (which we call \emph{Autoconj})
  operates directly on Python functions that compute log-joint
  distribution functions.  Autoconj provides support for
  conjugacy-exploiting algorithms in any Python-embedded PPL.  This
  paves the way for accelerating development of novel inference
  algorithms and structure-exploiting modeling strategies.%
  \setcounter{footnote}{0}%
  \footnote{
    Autoconj (including experiments) is available at
    \url{https://github.com/google-research/autoconj}.
  }
\end{abstract}

\section{Introduction}
\label{sec:intro}
Some models enjoy a property called \emph{conjugacy} that makes
computation easier. Conjugacy lets us compute \emph{complete
  conditional} distributions, that is, the distribution of some
variable conditioned on all other variables in the model. Complete
conditionals are at the heart of many classical statistical inference
algorithms such as Gibbs sampling \citep{geman1984stochastic},
coordinate-ascent variational inference \citep{jordan1999introduction}, and even
the venerable expectation-maximization (EM) algorithm
\citep{dempster1977maximum}. Conjugacy also makes it possible to marginalize
out some variables, which makes many algorithms faster and/or more
accurate \citep[e.g.; ][]{griffiths2004finding}. Many popular models in the
literature enjoy some form of conjugacy,
and these models can often be extended in
ways that preserve conjugacy.

For experienced researchers, deriving conditional and marginal
distributions using conjugacy relationships is straightforward. But it
is also time consuming and error prone, and diagnosing bugs in these
derivations can require significant effort \citep{cook2006validation}.

These considerations motivated specialized systems such as BUGS
\citep{spiegelhalter1995bugs}, VIBES \citep{winn2005variational}, and their many
successors. In these systems, one specifies a model in a probabilistic
programming language (PPL), provides observed values for some of the
model's variables, and lets the system automatically translate the
model specification into an algorithm (typically Gibbs sampling or
variational inference) that approximates the model's posterior
conditioned on the observed variables.

These systems are useful, but their monolithic design imposes a major
limitation: they are difficult to compose with other systems. For
example, a user who wants to interleave Gibbs sampling steps with some
customized Markov chain Monte Carlo (MCMC) kernel will find it very
difficult to take advantage of BUGS' Gibbs sampler.

In this paper, we propose a different strategy for exploiting
conditional conjugacy relationships. Unlike previous approaches (which
focus on relationships between pairs of random variables) our system
(which we call \emph{Autoconj}) operates directly on Python functions
that compute log-joint distribution functions. If asked to compute a
marginal distribution, Autoconj returns a Python function that
implements that marginal distribution's log-joint. If asked to compute
a complete conditional, it returns a Python function that returns
distribution objects.

Autoconj is not tied to
any particular approximate inference algorithm. But, because Autoconj is a
simple Python API, implementing conjugacy-exploiting approximate inference
algorithms using Autoconj is easy and fast (as we demonstrate in
section~\ref{sec:examples}). In particular, working in the Python/NumPy
ecosystem gives Autoconj users access to vectorized kernels, automatic
differentiation (via Autograd \citep{autograd}), sophisticated
optimization algorithms (via scipy.optimize), and even accelerated hardware
(via TensorFlow).

Autoconj provides
support for conjugacy-exploiting algorithms in any Python-embedded
PPL. More ambitiously, we hope that, just as automatic
differentiation has accelerated research in deep learning, Autoconj
will accelerate the development of novel inference algorithms and
modeling strategies that exploit conjugacy.

\section{Background: Exponential Families and Conjugacy}
\label{sec:bgnd}

To develop a system that can automatically find and exploit conjugacy,
we first develop a general perspective on exponential families.
Given a probability space~$(\X, \B(\X), \nu)$, where~$\B(\X)$ is the Borel
sigma algebra with respect to the standard topology on~$\X$, and a statistic
function~${t: \X \to \R^n}$, define the corresponding exponential family
of densities \citep{wainwright2008graphical}, indexed by the natural parameter $\eta \in
\R^n$, and log-normalizer function $\A$ as
\begin{equation}
  p(x \param \eta) = \exp \left\{ \langle \eta , \, t(x) \rangle - \A(\eta) \right\},
  \qquad
  \A(\eta) \triangleq \log \int \exp \left\{ \langle \eta, \, t(x) \rangle \right\} \, \nu(\diff x),
  \label{expfam}
\end{equation}
where $\langle \cdot \, , \, \cdot \rangle$ denotes the standard inner product.
The log-normalizer function $\A$ is directly related to the cumulant-generating
function, and in particular it satisfies
\begin{equation}
  \nabla \A(\eta) = \E \left[ t(x) \right],
  \qquad
  \nabla^2 \A(\eta) = \E \left[ t(x) t(x)^\T \right] - \E \left[ t(x) \right] \E \left[ t(x) \right]^\T,
\end{equation}
where the expectation is with respect to~$p(x \param \eta)$.
For a given statistic function $t$, when the corresponding distribution can be
sampled efficiently, and when $\A$ and its derivatives can be evaluated efficiently,
we say the exponential family (or the statistic function that defines it) is
\emph{tractable}.

Consider an exponential-family model where the log density has the form
\begin{align}
  \log p(\bz, x)
  = \log p(z_1, z_2, \ldots, z_M, x)
  &=\textstyle \sum_{\beta \in \boldsymbol{\beta}}
  \langle \eta_\beta(x), \; t_{\z_1}(z_1)^{\beta_1} \otimes \cdots \otimes t_{\z_M}(z_M)^{\beta_M} \rangle
  \notag
  \\
  &\triangleq g(t_{\z_1}(z_1), \ldots, t_{\z_M}(z_M)),
  \label{eq:multilinear energy}
\end{align}
where $\bbeta \subseteq \{0,1\}^M$ is an index set, we take $t_{\z_m}(z_m)^0
\equiv 1$, and where the functions $\{t_{\z_m}(z_m)\}_{m=1}^M$ are each the
sufficient statistics of a tractable exponential family.
In words, the log joint density $\log p(\bz, x)$ can be written as a multilinear (or
multiaffine) polynomial $g$ applied to the statistic functions
$\{t_{\z_m}(z_m)\}$.
These models arise when building complex distributions from simpler, tractable
ones, and the algebraic structure in $g$ corresponds to graphical model
structure \citep{wainwright2008graphical,koller2009probabilistic}.
In general the posterior $\log p(\bz \given x)$ is not tractable, but it
admits efficient approximate inference algorithms.

Models of the form~\eqref{eq:multilinear energy} are
known as \emph{conditionally conjugate} models. Each conditional
$p(z_m \given z_{\neg m})$ (where $z_{\neg m} \triangleq \{z_1, \ldots, z_M \} \setminus
\{z_m\}$) is a tractable exponential family.
Moreover, the parameters of these conditional densities can be extracted
using differentiation. We formalize this below.

\begin{claim}
    Given an exponential family with density of the form~\eqref{eq:multilinear energy}, we have
    \begin{equation}
        p(z_m \given z_{\neg m})
        = \exp \left\{ \langle \eta_{\z_m}^*, \, t_{\z_m}(z_m) \rangle
        - \A_{\z_m}(\eta_{\z_m}^*) \right\}
        \;\,
        \text{where}
        \;\,
        \eta_{\z_m}^* \triangleq \nabla_{t_{z_m}} g(t_{\z_1}(z_1), \ldots, t_{\z_M}(z_M)).
        \notag
    \end{equation}
\end{claim}

As a consequence, if we had code for evaluating the functions $g$ and
$\{t_{\z_m}\}$, along with a table of sampling routines corresponding to each
tractable statistic $t_{\z_m}$, then we could use automatic differentiation to write
a generic Gibbs sampling algorithm.
This generic algorithm could be extended to work with any tractable
exponential-family distribution simply by populating a table matching tractable statistics
functions to their corresponding samplers.
Note this differs from a table of pairs of random variables:
conjugacy derives from this lower-level algebraic relationship.

The model structure~\eqref{eq:multilinear energy} can be exploited in other
approximate inference algorithms, including variational mean field
\citep{wainwright2008graphical} and stochastic variational inference
\citep{hoffman2013stochastic}.
Consider the variational distribution
\begin{equation}
  q(z) = \prod_m q(z_m \param \eta_{\z_m}),
  \qquad
  q(z_m \param \eta_{\z_m}) = \exp \left\{ \langle \eta_{\z_m}, \, t_{\z_m}(z_m) \rangle - \A_{\z_m}(\eta_{\z_m}) \right\},
  \label{eq:variational factorization}
\end{equation}
where~$\eta_{\z_m}$ are natural parameters of the variational factors.
We write the variational evidence lower bound objective $\L =
\L(\eta_{\z_1}, \ldots, \eta_{\z_M})$ for approximating the posterior $p(z \given x)$ as
\begin{equation}
  \log p(x) = \log \int p(\bz, x) \, \nu_\z(\mathrm{d}\bz)
  = \log \E_{q(\bz)} \! \! \left[ \frac{p(\bz, x)}{q(\bz)} \right] \!
  \geq \E_{q(\bz)} \! \! \left[ \log \frac{p(\bz, x)}{q(\bz)} \right] \!
  \triangleq \L.
  \label{eq:variational objective}
\end{equation}
We can write block coordinate ascent updates for this objective using
differentiation:

\begin{claim}
    Given a model with density of the form~\eqref{eq:multilinear energy} and
    variational problem~\eqref{eq:variational
    factorization}-\eqref{eq:variational objective}, we have
    \begin{equation}
        \argmax_{\eta_{\z_m}} \L(\eta_{\z_1}, \ldots \eta_{\z_M})
        = \nabla_{\mu_{\z_m}} g(\mu_{\z_1}, \ldots, \mu_{\z_M})
        \;\,
        \text{where}
        \;\,
        \mu_{\z_{m'}} \triangleq \nabla \A_{\z_{m'}}(\eta_{\z_{m'}}),
        \; m'=1,\ldots,M.
        \notag
    \end{equation}
\end{claim}

Thus if we had code for evaluating the functions $g$ and $\{t_{\z_m}\}$, along
with a table of log-normalizer functions $\A_{\z_m}$ corresponding to each
tractable statistic $t_{\z_m}$, then we could use automatic differentiation to
write a generic block coordinate-ascent variational inference algorithm.
New tractable structures could be added to this algorithm's repertoire simply
by populating the table of statistics and their corresponding log-normalizers.

If all this tractable exponential-family structure can be exploited
generically, why is writing conjugacy-exploiting inference software still so
laborious?
The reason is that it is not always easy to get our hands on the
representation~\eqref{eq:multilinear energy}.
Even when a model's log joint density \emph{could} be written as
in~\eqref{eq:multilinear energy}, it is often difficult and error-prone to write
code to evaluate $g$ directly; it is much more natural to specify model
densities without being constrained to this form.
The situation is analogous to deep learning research before flexible automatic
differentiation: we're stuck writing too much code by hand, and even though in
principle this process could be automated, our current software tools aren't up
to the task unless we're willing to get locked into a limited mini-language.

Based on this derivation, Autoconj is built to automatically extract these tractable
structures (i.e., the functions $g$ and $\{t_{\z_m}\}$). It does this
given log density functions written in plain Python and NumPy. And it
reaps automatic structure-exploiting inference algorithms as a result.

\section{Analyzing Log-Joint Functions}

To extract sufficient statistics and natural parameters from a
log-joint function, Autoconj first represent that function in a convenient
canonical form.
It applies a canonicalization process, which comprises two stages: 1. a tracer
maps Python log-joint probability functions to symbolic term graphs; 2. a
domain-specific rewrite system puts the log-joint functions in a
canonical form and extracts the component functions defined in
Section~\ref{sec:bgnd}.

\begin{figure}[!tb]
  \centering
  \begin{subfigure}[t]{0.35\columnwidth}
    \centering
    \includegraphics[width=\columnwidth]{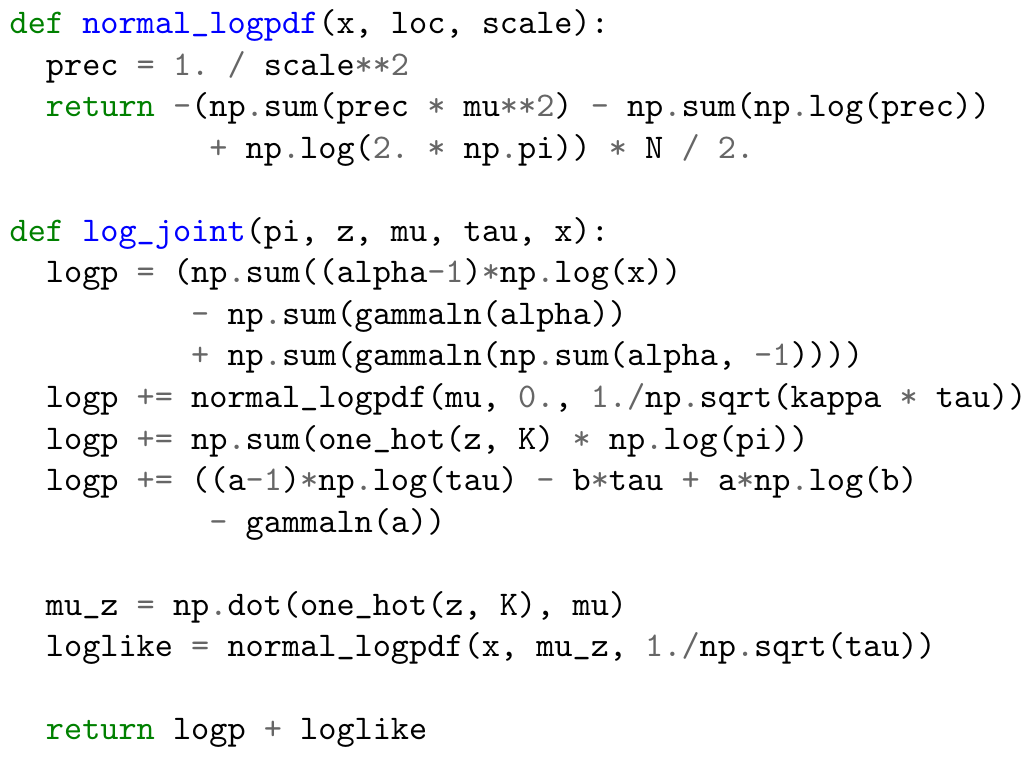}
  \end{subfigure}
  \hskip -0.05\columnwidth
  \begin{subfigure}[t]{0.69\columnwidth}
    \centering
   \includegraphics[width=\columnwidth]{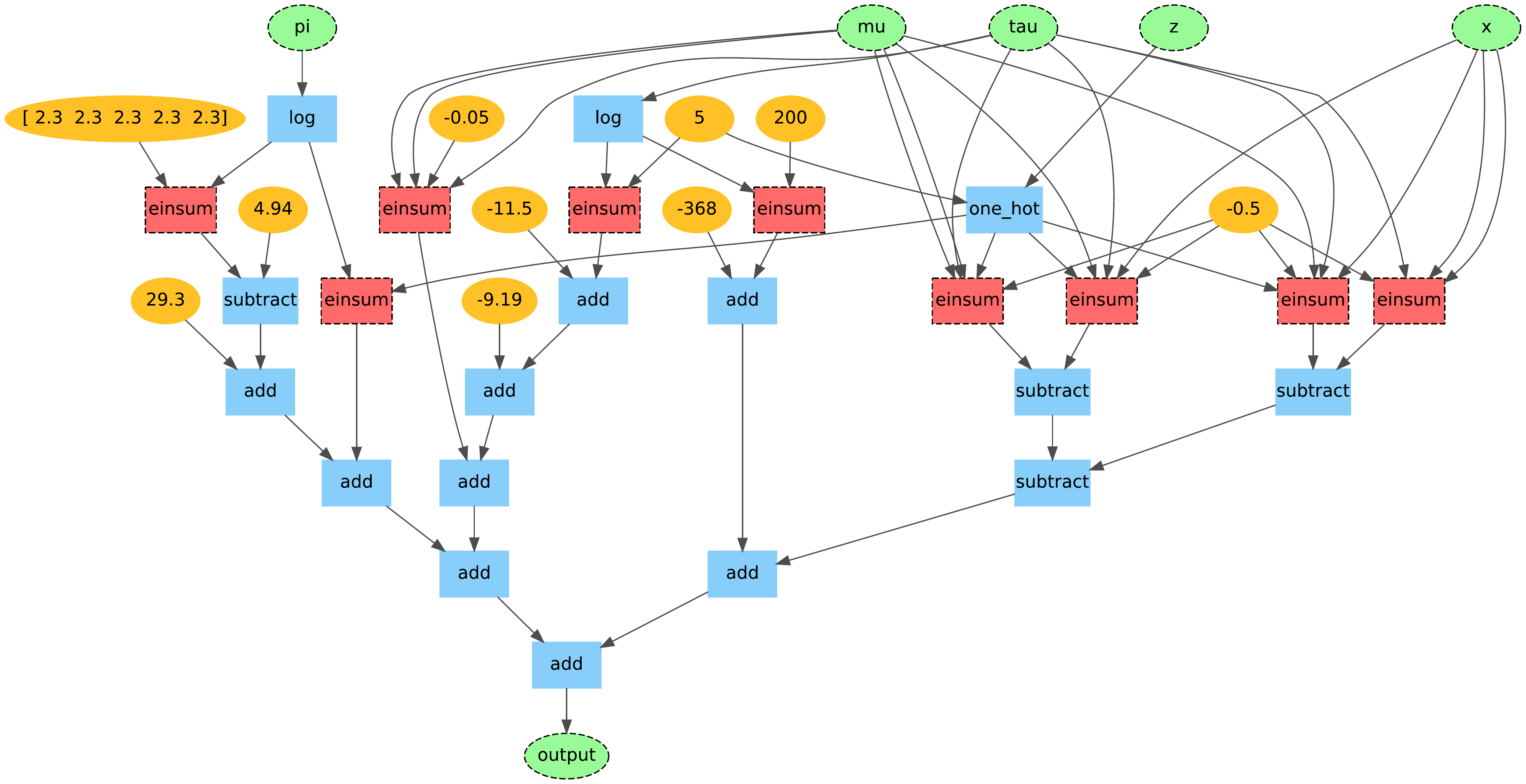}
  \end{subfigure}
  \caption{
    Left: Python code for evaluating the log joint density of a Gaussian
    mixture model.
    Right: canonicalized computation graph, representing the same log joint
    density function but rewritten as a sum of \texttt{np.einsum}s of statistic
    functions.
  }
  \label{fig:canonicalization}
\end{figure}

\subsection{Tracing Python programs to generate term graphs}

The tracer's purpose is to map a Python function denoting a log-joint function
to an acyclic term graph data structure.
It accomplishes this mapping without having to analyze Python syntax or
reason about its semantics directly; instead, the tracer monitors the execution
of a Python function in terms of the primitive functions that are applied to its
arguments to produce its final output.
As a consequence, intermediates like non-primitive function calls and auxiliary
data structures, including tuples/lists/dicts as well as custom classes, do
not appear in the trace and instead all get traced through.
The ultimate output of the tracer is a directed acyclic data flow graph, where
nodes represent application of primitive functions (typically NumPy functions)
and edges represent data flow.
This approach is both simple to implement and able to handle
essentially any Python code.

A weakness of this tracing approach is that we only trace one evaluation of the
function on example arguments, and we assume that the trace represents the
same mathematical function that the original Python code denotes.
This assumption can fail.
For example, if a Python function has an if/else that depends on the value of
the arguments (and is not expressed in a primitive function), then the tracer
could only follow one branch, and so instead raises an error.
In the context of tracing log-joint functions, this limitation does not seem
to arise too frequently, but it does affect our handling of discrete random
variables; for densities of discrete random variables, the tracer can intercept
either indexing expressions like \texttt{pi[z]} or the use of the primitive
function \texttt{one\_hot}.

Figure~\ref{fig:canonicalization} summarizes the tracer's use on Python code to generate a term graph.
To implement the tracing mechanism, we reuse Autograd's tracer
\citep{autograd}, which is designed to be general-purpose and
extensible with a simple API.
Other similar tracing mechanisms are common in probabilistic programming
\citep{dippl}.

\subsection{Domain-specific term graph rewriting system}

The goal of the rewrite system is to take a log-joint term graph and
manipulate it into a canonical form.
Mathematically, the canonical form described in Section~\ref{sec:bgnd} is a
multilinear polynomial $g$ on tensor-valued statistic functions $t_1, \ldots,
t_M$.
For term graphs, we say a term graph is in this canonical form when its output
node represents a sum of \texttt{np.einsum} nodes, with each \texttt{np.einsum}
node corresponding to a monomial term in $g$ and each \texttt{np.einsum}
argument being either a constant, a nonlinear function of an input, or an
input itself, with the latter two cases corresponding to statistic functions
$t_m$.
We rely on \texttt{np.einsum} because it is capable of expressing arbitrary
tensor contractions, meaning it is a uniform way to express arbitrary monomial
terms in $g$.

At its core, the rewrite system is based on pattern-directed invocation of
rewrite rules, each of which can match and then modify a small subgraph
corresponding to a few primitive function applications.
Our pattern language is a new Python-embedded DSL, which is compiled into
continuation-passing matcher combinators \citep{rules}.
In addition to basic matchers for data types and each primitive function, the
pattern combinators include \texttt{Choice}, which produces a match if any of
its argument combinators produce a match, and \texttt{Segment}, which can match
any number of elements in a list, including argument lists.
By using continuation passing, backtracking is effectively handled by the
Python call stack, and it's straightforward to extract just one match or all
possible matches.
The pattern language compiler is only \textasciitilde 300 lines and is fully
extensible by registering new syntax handlers.

A rewrite rule is then a pattern paired with a rewriter function.
A rewriter essentially represents a syntactic macro operating on the term
subgraph, using matched sub-terms collected by the pattern to generate a new
term subgraph.
To specify each rewriter, we again make use of the tracer: we simply write a
Python function that, when traced on appropriate arguments, produces the new
subgraph, which we then patch into the term graph.
This mechanism is analogous to quasiquoting \citep{rules}, since it specifies a
syntactic transformation in terms of native Python expressions.
Thus by using pattern matching and tracing-based rewriters, we can define
general rewrite rules without writing any code that manually traverses or
modifies the term graph data structure.
As a result, it is straightforward to add new rewrite rules to the system.
See Listing~\ref{code:rewrite rule} for an example rewrite rule.

\begin{listing}[htb]


\begin{Verbatim}[commandchars=\\\{\},codes={\catcode`\$=3\catcode`\^=7\catcode`\_=8}]
\PYG{n}{pat} \PYG{o}{=} \PYG{p}{(}\PYG{n}{Einsum}\PYG{p}{,} \PYG{n}{Str}\PYG{p}{(}\PYG{l+s+s1}{\PYGZsq{}formula\PYGZsq{}}\PYG{p}{),} \PYG{n}{Segment}\PYG{p}{(}\PYG{l+s+s1}{\PYGZsq{}args1\PYGZsq{}}\PYG{p}{),}
       \PYG{p}{(}\PYG{n}{Choice}\PYG{p}{(}\PYG{n}{Subtract}\PYG{p}{(}\PYG{l+s+s1}{\PYGZsq{}op\PYGZsq{}}\PYG{p}{),} \PYG{n}{Add}\PYG{p}{(}\PYG{l+s+s1}{\PYGZsq{}op\PYGZsq{}}\PYG{p}{)),} \PYG{n}{Val}\PYG{p}{(}\PYG{l+s+s1}{\PYGZsq{}x\PYGZsq{}}\PYG{p}{),} \PYG{n}{Val}\PYG{p}{(}\PYG{l+s+s1}{\PYGZsq{}y\PYGZsq{}}\PYG{p}{)),} \PYG{n}{Segment}\PYG{p}{(}\PYG{l+s+s1}{\PYGZsq{}args2\PYGZsq{}}\PYG{p}{))}

\PYG{k}{def} \PYG{n+nf}{rewriter}\PYG{p}{(}\PYG{n}{formula}\PYG{p}{,} \PYG{n}{op}\PYG{p}{,} \PYG{n}{x}\PYG{p}{,} \PYG{n}{y}\PYG{p}{,} \PYG{n}{args1}\PYG{p}{,} \PYG{n}{args2}\PYG{p}{):}
  \PYG{k}{return} \PYG{n}{op}\PYG{p}{(}\PYG{n}{np}\PYG{o}{.}\PYG{n}{einsum}\PYG{p}{(}\PYG{n}{formula}\PYG{p}{,} \PYG{o}{*}\PYG{p}{(}\PYG{n}{args1} \PYG{o}{+} \PYG{p}{(}\PYG{n}{x}\PYG{p}{,)} \PYG{o}{+} \PYG{n}{args2}\PYG{p}{)),}
            \PYG{n}{np}\PYG{o}{.}\PYG{n}{einsum}\PYG{p}{(}\PYG{n}{formula}\PYG{p}{,} \PYG{o}{*}\PYG{p}{(}\PYG{n}{args1} \PYG{o}{+} \PYG{p}{(}\PYG{n}{y}\PYG{p}{,)} \PYG{o}{+} \PYG{n}{args2}\PYG{p}{)))}

\PYG{n}{distribute\PYGZus{}einsum} \PYG{o}{=} \PYG{n}{Rule}\PYG{p}{(}\PYG{n}{pat}\PYG{p}{,} \PYG{n}{rewriter}\PYG{p}{)}  \PYG{c+c1}{\PYGZsh{} Rule is a namedtuple}
\end{Verbatim}
\caption{A rewrite for distributing \texttt{np.einsum} over addition and
subtraction.}
\label{code:rewrite rule}
\end{listing}

Rewrite rules are composed into a term rewriting system by an alternating
strategy with two steps.
In the first step, for each rule we look for a pattern match anywhere in the
term graph starting from the output; if no match is found then the process
terminates, and if there is a match we apply the corresponding rewriter and
move to the second step.
In the second step, we traverse the graph from the inputs to the output,
performing common subexpression elimination (CSE) and applying local
simplifications that only involve one primitive at a time (like replacing a
\texttt{np.dot} with an equivalent \texttt{np.einsum}) and hence don't require
pattern matching.
By alternating rewrites with CSE, we remove any redundancies introduced by the
rewrites.
It is straightforward to compose new rewrite systems, involving different sets
of rewrite rules or different strategies for applying them.

The process is summarized in Figure~\ref{fig:canonicalization}.
The rewriting process aims to transform the term graph of a log joint density
into the canonical sum-of-einsums polynomial form corresponding to
Eq.~\eqref{eq:multilinear energy} (up to commutativity).
We do not have a proof that the rewrites are terminating or
confluent~\citep{baader1999term}, and the set of possible terms is very
complex, though intuitively each rewrite rule applied makes strict progress
towards the canonical form (e.g.~by distributing multiplication across
addition).
In practice there have been no problems with termination or normalization.

Once we have processed the log-joint term graph into a canonical
form, it is straightforward to extract the objects of interest (namely the
statistic functions $t_1, \ldots, t_M$ and the polynomial $g$), match the
tractable statistics with corresponding log-normalizer and sampler functions
from a table, and perform any further manipulations like automatic
differentiation.
Moreover, we can map the term graph back into a Python function (via an
interpreter), so the rewrite system is hermetic: we can use its output with any
other Python tools, like Autograd or SciPy, without those tools needing to know
anything about it.

Term rewriting systems have a long history in compilers and symbolic math
systems \citep{scmutils,rules,pyrewrite,rozenberg1997handbook,baader1999term}.
The main novelty here is the application domain and specific concerns and
capabilities that arise from it; we're manipulating exponential families of
densities for multidimensional random variables, and hence our system is
focused on matrix and tensor manipulations, which have limited support in other
systems, and a specific canonical form informed by structure-exploiting
approximate inference algorithms.
Our implementation is closely related to the term rewriting system in
\texttt{scmutils} \citep{scmutils} and Rules \citep{rules}, which also use a pattern
language (embedded in Scheme) based on continuation-passing matcher combinators
and quasiquote-based syntactic macros.
Two differences in the implementation are that our system operates on term
graphs rather than syntax trees, and that we use tracing to implement a kind of
macro system on our term graph data structures (instead of using Scheme's
built-in quasiquotes and homoiconicity).

\subsection{Recognizing Sufficient Statistics and Natural Parameters}
Once the log-joint graph has been canonicalized as a sum of
\texttt{np.einsum}s of functions of the inputs, we can discover and
extract exponential-family structure.

Suppose we are interested in the complete conditional of an input $z$.
We first need to find all nodes that represent sufficient statistics
of $z$. We begin at the output node, and search up through the graph,
ignoring any nodes that do not depend on $z$. We walk through any
\texttt{add} or \texttt{subtract} nodes until we reach an
\texttt{np.einsum} node.  If $z$ is a parent of more than one argument
of that \texttt{np.einsum} node, then the node represents a nonlinear
function of $z$ and we label it as a sufficient statistic (if the node
has any inputs that do not depend on $z$ we also need to split those
out). Otherwise, we walk through the \texttt{np.einsum} node since it
is a linear function of $z$. If at any point in the search we reach
either $z$ or a node that is not linear in $z$ (i.e., an \texttt{add},
\texttt{subtract}, or \texttt{np.einsum}), we label it as a sufficient
statistic.

Once we have found the set of sufficient statistic nodes, we can
determine whether they correspond to a known tractable exponential
family. For example, in Figure~\ref{fig:canonicalization}, $z$ has
integer support and the one-hot statistic, so its complete conditional
is a categorical distribution; $\pi$'s support is the simplex and its
only sufficient statistic is $\log \pi$, so $\pi$'s complete
conditional is a Dirichlet; $\tau$'s support is the non-negative
reals, and its sufficient statistics are $\tau$ and $\log\tau$, so its
complete conditional is a gamma distribution. If the
sufficient-statistic functions do not correspond to a known exponential
family, then the system raises an exception.

Finally, to get the natural parameters we can simply take the symbolic
gradient of the output node with respect to each sufficient-statistic
node using Autograd.

\section{Related Work}
Many probabilistic programming languages (PPLs) exploit conjugacy
relationships. PPLs like BUGS \citep{spiegelhalter1995bugs},
VIBES \citep{winn2005variational}, and
Augur \citep{tristan2014augur}
build an explicit graph of
random variables and find conjugate pairs in that graph.
This
strategy remains widely applicable, but ties the system very strongly
to the PPL's model representation.
Most recently, Birch \citep{murray2018delayed} utilizes a flexible strategy for
combining conjugacy and approximate inference in order to enable
algorithms such as Sequential Monte Carlo with Rao-Blackwellization.
Autoconj could extend their conjugacy component.

PPLs such as Hakaru \citep{narayanan2016hakaru}
have considered treating conditioning and marginalization as program
transformations based on computer algebra
\citep{carette2016algebra,gehr2016psi}. Unfortunately, most existing
computer algebra systems have very limited support for linear algebra
and multidimensional array processing, which in turn makes it hard for
these systems to either express models using NumPy-style
broadcasting or take advantage of vectorized hardware (although
\citet{narayanan2017symbolic} take steps to address this).  Exploiting
multivariate-Gaussian structure in these languages is particularly
cumbersome. Orthogonal to our work, \citet{narayanan2017symbolic}
advances symbolic manipulation for general probability spaces such as mixed
discrete-and-continuous events. These ideas could also be used in Autoconj.

\section{Examples and Experiments}
\label{sec:examples}
In this section we provide code snippets and empirical
results to demonstrate Autoconj's functionality, as well as the
benefits of being embedded in Python as opposed to a more narrowly
focused domain-specific language. We begin with some examples.

Listing~\ref{code:beta-bernoulli} demonstrates doing
exact conditioning and marginalization in a trivial Beta-Bernoulli
model. The log-joint is implemented using NumPy, and is passed to
\texttt{complete\_conditional()} and \texttt{marginalize()}. These
functions also take an \texttt{argnum} parameter that says which parameter
to marginalize out or take the complete conditional of (0 in this example,
referring to \texttt{counts_prob}) and a \texttt{support} parameter.
Finally, they take a list of dummy arguments that are used to propagate shapes
and types when tracing the log-joint function.
\begin{listing}[htb]

\begin{Verbatim}[commandchars=\\\{\},codes={\catcode`\$=3\catcode`\^=7\catcode`\_=8}]
\PYG{k}{def} \PYG{n+nf}{log\PYGZus{}joint}\PYG{p}{(}\PYG{n}{counts\PYGZus{}prob}\PYG{p}{,} \PYG{n}{n\PYGZus{}heads}\PYG{p}{,} \PYG{n}{n\PYGZus{}draws}\PYG{p}{,} \PYG{n}{prior\PYGZus{}a}\PYG{p}{,} \PYG{n}{prior\PYGZus{}b}\PYG{p}{):}
  \PYG{n}{log\PYGZus{}prob} \PYG{o}{=} \PYG{p}{(}\PYG{n}{prior\PYGZus{}a}\PYG{o}{\PYGZhy{}}\PYG{l+m+mi}{1}\PYG{p}{)}\PYG{o}{*}\PYG{n}{np}\PYG{o}{.}\PYG{n}{log}\PYG{p}{(}\PYG{n}{counts\PYGZus{}prob}\PYG{p}{)} \PYG{o}{+} \PYG{p}{(}\PYG{n}{prior\PYGZus{}b}\PYG{o}{\PYGZhy{}}\PYG{l+m+mi}{1}\PYG{p}{)}\PYG{o}{*}\PYG{n}{np}\PYG{o}{.}\PYG{n}{log1p}\PYG{p}{(}\PYG{o}{\PYGZhy{}}\PYG{n}{counts\PYGZus{}prob}\PYG{p}{)}
  \PYG{n}{log\PYGZus{}prob} \PYG{o}{+=} \PYG{n}{n\PYGZus{}heads}\PYG{o}{*}\PYG{n}{np}\PYG{o}{.}\PYG{n}{log}\PYG{p}{(}\PYG{n}{counts\PYGZus{}prob}\PYG{p}{)} \PYG{o}{+} \PYG{p}{(}\PYG{n}{n\PYGZus{}draws}\PYG{o}{\PYGZhy{}}\PYG{n}{n\PYGZus{}heads}\PYG{p}{)}\PYG{o}{*}\PYG{n}{np}\PYG{o}{.}\PYG{n}{log1p}\PYG{p}{(}\PYG{o}{\PYGZhy{}}\PYG{n}{counts\PYGZus{}prob}\PYG{p}{)}
  \PYG{n}{log\PYGZus{}prob} \PYG{o}{+=} \PYG{o}{\PYGZhy{}}\PYG{n}{gammaln}\PYG{p}{(}\PYG{n}{prior\PYGZus{}a}\PYG{p}{)} \PYG{o}{\PYGZhy{}} \PYG{n}{gammaln}\PYG{p}{(}\PYG{n}{prior\PYGZus{}b}\PYG{p}{)} \PYG{o}{+} \PYG{n}{gammaln}\PYG{p}{(}\PYG{n}{prior\PYGZus{}a} \PYG{o}{+} \PYG{n}{prior\PYGZus{}b}\PYG{p}{)}
  \PYG{k}{return} \PYG{n}{log\PYGZus{}prob}

\PYG{n}{n\PYGZus{}heads}\PYG{p}{,} \PYG{n}{n\PYGZus{}draws} \PYG{o}{=} \PYG{l+m+mi}{60}\PYG{p}{,} \PYG{l+m+mi}{100}
\PYG{n}{prior\PYGZus{}a}\PYG{p}{,} \PYG{n}{prior\PYGZus{}b} \PYG{o}{=} \PYG{l+m+mf}{0.5}\PYG{p}{,} \PYG{l+m+mf}{0.5}
\PYG{n}{all\PYGZus{}args} \PYG{o}{=} \PYG{p}{[}\PYG{l+m+mf}{0.5}\PYG{p}{,} \PYG{n}{n\PYGZus{}heads}\PYG{p}{,} \PYG{n}{n\PYGZus{}draws}\PYG{p}{,} \PYG{n}{prior\PYGZus{}a}\PYG{p}{,} \PYG{n}{prior\PYGZus{}b}\PYG{p}{]}
\PYG{n}{make\PYGZus{}complete\PYGZus{}conditional} \PYG{o}{=} \PYG{n}{autoconj}\PYG{o}{.}\PYG{n}{complete\PYGZus{}conditional}\PYG{p}{(}
    \PYG{n}{log\PYGZus{}joint}\PYG{p}{,} \PYG{l+m+mi}{0}\PYG{p}{,} \PYG{n}{SupportTypes}\PYG{o}{.}\PYG{n}{UNIT\PYGZus{}INTERVAL}\PYG{p}{,} \PYG{o}{*}\PYG{n}{all\PYGZus{}args}\PYG{p}{)}
\PYG{c+c1}{\PYGZsh{} A Beta(60.5, 40.5) distribution object.}
\PYG{n}{complete\PYGZus{}conditional} \PYG{o}{=} \PYG{n}{make\PYGZus{}complete\PYGZus{}conditional}\PYG{p}{(}\PYG{n}{n\PYGZus{}heads}\PYG{p}{,} \PYG{n}{n\PYGZus{}draws}\PYG{p}{,} \PYG{n}{prior\PYGZus{}a}\PYG{p}{,} \PYG{n}{prior\PYGZus{}b}\PYG{p}{)}
\PYG{c+c1}{\PYGZsh{} Computes the marginal log\PYGZhy{}probability of n\PYGZus{}heads, n\PYGZus{}draws given prior\PYGZus{}a, prior\PYGZus{}b}
\PYG{n}{marginal} \PYG{o}{=} \PYG{n}{autoconj}\PYG{o}{.}\PYG{n}{marginalize}\PYG{p}{(}\PYG{n}{log\PYGZus{}joint}\PYG{p}{,} \PYG{l+m+mi}{0}\PYG{p}{,} \PYG{n}{SupportTypes}\PYG{o}{.}\PYG{n}{UNIT\PYGZus{}INTERVAL}\PYG{p}{,} \PYG{o}{*}\PYG{n}{all\PYGZus{}args}\PYG{p}{)}
\PYG{k}{print}\PYG{p}{(}\PYG{l+s+s1}{\PYGZsq{}log p(n\PYGZus{}heads=60 | a, b) =\PYGZsq{}}\PYG{p}{,} \PYG{n}{marginal}\PYG{p}{(}\PYG{n}{n\PYGZus{}heads}\PYG{p}{,} \PYG{n}{n\PYGZus{}draws}\PYG{p}{,} \PYG{n}{prior\PYGZus{}a}\PYG{p}{,} \PYG{n}{prior\PYGZus{}b}\PYG{p}{))}
\end{Verbatim}
\caption{Exact inference in a simple Beta-Bernoulli model.}
\label{code:beta-bernoulli}
\end{listing}

Listing~\ref{code:normal-gamma} demonstrates how one can handle a more
complicated compound prior: the normal-gamma distribution, which is
the natural conjugate prior for Bayesian linear regression. Note that
we can call \texttt{complete\_conditional()} on the function produced by
\texttt{marginalize()}.
\begin{listing}[htb]


\begin{Verbatim}[commandchars=\\\{\},codes={\catcode`\$=3\catcode`\^=7\catcode`\_=8}]
\PYG{k}{def} \PYG{n+nf}{log\PYGZus{}joint}\PYG{p}{(}\PYG{n}{tau}\PYG{p}{,} \PYG{n}{beta}\PYG{p}{,} \PYG{n}{x}\PYG{p}{,} \PYG{n}{y}\PYG{p}{,} \PYG{n}{a}\PYG{p}{,} \PYG{n}{b}\PYG{p}{,} \PYG{n}{kappa}\PYG{p}{,} \PYG{n}{mu0}\PYG{p}{):}
  \PYG{n}{log\PYGZus{}p\PYGZus{}tau} \PYG{o}{=} \PYG{n}{log\PYGZus{}probs}\PYG{o}{.}\PYG{n}{gamma\PYGZus{}gen\PYGZus{}log\PYGZus{}prob}\PYG{p}{(}\PYG{n}{tau}\PYG{p}{,} \PYG{n}{a}\PYG{p}{,} \PYG{n}{b}\PYG{p}{)}
  \PYG{n}{log\PYGZus{}p\PYGZus{}beta} \PYG{o}{=} \PYG{n}{log\PYGZus{}probs}\PYG{o}{.}\PYG{n}{norm\PYGZus{}gen\PYGZus{}log\PYGZus{}prob}\PYG{p}{(}\PYG{n}{beta}\PYG{p}{,} \PYG{n}{mu0}\PYG{p}{,} \PYG{l+m+mf}{1.} \PYG{o}{/} \PYG{n}{np}\PYG{o}{.}\PYG{n}{sqrt}\PYG{p}{(}\PYG{n}{kappa} \PYG{o}{*} \PYG{n}{tau}\PYG{p}{))}
  \PYG{n}{log\PYGZus{}p\PYGZus{}y} \PYG{o}{=} \PYG{n}{log\PYGZus{}probs}\PYG{o}{.}\PYG{n}{norm\PYGZus{}gen\PYGZus{}log\PYGZus{}prob}\PYG{p}{(}\PYG{n}{y}\PYG{p}{,} \PYG{n}{np}\PYG{o}{.}\PYG{n}{dot}\PYG{p}{(}\PYG{n}{x}\PYG{p}{,} \PYG{n}{beta}\PYG{p}{),} \PYG{l+m+mf}{1.} \PYG{o}{/} \PYG{n}{np}\PYG{o}{.}\PYG{n}{sqrt}\PYG{p}{(}\PYG{n}{tau}\PYG{p}{))}
  \PYG{k}{return} \PYG{n}{log\PYGZus{}p\PYGZus{}tau} \PYG{o}{+} \PYG{n}{log\PYGZus{}p\PYGZus{}beta} \PYG{o}{+} \PYG{n}{log\PYGZus{}p\PYGZus{}y}

\PYG{c+c1}{\PYGZsh{} log p(tau, x, y), marginalizing out beta}
\PYG{n}{tau\PYGZus{}x\PYGZus{}y\PYGZus{}log\PYGZus{}prob} \PYG{o}{=} \PYG{n}{autoconj}\PYG{o}{.}\PYG{n}{marginalize}\PYG{p}{(}\PYG{n}{log\PYGZus{}joint}\PYG{p}{,} \PYG{l+m+mi}{1}\PYG{p}{,} \PYG{n}{SupportTypes}\PYG{o}{.}\PYG{n}{REAL}\PYG{p}{,} \PYG{o}{*}\PYG{n}{all\PYGZus{}args}\PYG{p}{)}
\PYG{c+c1}{\PYGZsh{} compute and sample from p(tau | x, y)}
\PYG{n}{make\PYGZus{}tau\PYGZus{}posterior} \PYG{o}{=} \PYG{n}{autoconj}\PYG{o}{.}\PYG{n}{complete\PYGZus{}conditional}\PYG{p}{(}
    \PYG{n}{tau\PYGZus{}x\PYGZus{}y\PYGZus{}log\PYGZus{}prob}\PYG{p}{,} \PYG{l+m+mi}{0}\PYG{p}{,} \PYG{n}{SupportTypes}\PYG{o}{.}\PYG{n}{NONNEGATIVE}\PYG{p}{,} \PYG{o}{*}\PYG{n}{all\PYGZus{}args\PYGZus{}ex\PYGZus{}beta}\PYG{p}{)}
\PYG{n}{tau\PYGZus{}sample} \PYG{o}{=} \PYG{n}{make\PYGZus{}tau\PYGZus{}posterior}\PYG{p}{(}\PYG{n}{x}\PYG{p}{,} \PYG{n}{y}\PYG{p}{,} \PYG{n}{a}\PYG{p}{,} \PYG{n}{b}\PYG{p}{,} \PYG{n}{kappa}\PYG{p}{,} \PYG{n}{mu0}\PYG{p}{)}\PYG{o}{.}\PYG{n}{rvs}\PYG{p}{()}
\PYG{c+c1}{\PYGZsh{} compute and sample from p(beta | tau, x, y)}
\PYG{n}{make\PYGZus{}beta\PYGZus{}conditional} \PYG{o}{=} \PYG{n}{autoconj}\PYG{o}{.}\PYG{n}{complete\PYGZus{}conditional}\PYG{p}{(}
    \PYG{n}{log\PYGZus{}joint}\PYG{p}{,} \PYG{l+m+mi}{1}\PYG{p}{,} \PYG{n}{SupportTypes}\PYG{o}{.}\PYG{n}{REAL}\PYG{p}{,} \PYG{o}{*}\PYG{n}{all\PYGZus{}args}\PYG{p}{)}
\PYG{n}{beta\PYGZus{}sample} \PYG{o}{=} \PYG{n}{make\PYGZus{}beta\PYGZus{}conditional}\PYG{p}{(}\PYG{n}{tau}\PYG{p}{,} \PYG{n}{x}\PYG{p}{,} \PYG{n}{y}\PYG{p}{,} \PYG{n}{a}\PYG{p}{,} \PYG{n}{b}\PYG{p}{,} \PYG{n}{kappa}\PYG{p}{,} \PYG{n}{mu0}\PYG{p}{)}
\end{Verbatim}
\caption{Exact inference in a Bayesian linear regression with normal-gamma
    compound prior. We factorize the joint posterior on the mean and precision
    as $p(\mu, \tau\mid x, y) = p(\tau\mid x, y) p(\mu\mid x, y, \tau)$. We
    first compute the \emph{marginal} joint distribution $p(x, y, \tau)$ by
    calling \texttt{marginalize()} on the full log-joint. We then compute the
    marginal posterior $p(\tau\mid x, y)$ by calling
    \texttt{complete\_conditional()} on the marginal $p(x, y, \tau)$, and
    finally we compute $p(\mu\mid x, y, \tau)$ by calling
\texttt{complete\_conditional()} on the full log-joint.}
\label{code:normal-gamma}
\end{listing}

We can extend the marginalize-and-condition strategy above to more
complicated models.  In the supplement, we demonstrate how one can
implement the Kalman-filter recursion with Autoconj. The generative
process is
\begin{equation}
\begin{split}
x_1\sim \operatorname{Normal}(0, s_0);\quad
x_{t>1}\sim \operatorname{Normal}(x_{t-1}, s_x);\quad
y_t\sim \operatorname{Normal}(x_{t}, s_y).
\end{split}
\end{equation}
The core recursion consists of using \texttt{marginalize()} to compute
$p(x_{t+1}, y_{t+1} \mid y_{1:t})$ from the functions $p(x_t \mid
y_{1:t})$ and $p(x_{t+1}, y_{t+1}\mid x_t)$, then using
\texttt{marginalize()} again to compute $p(y_{t+1}\mid y_{1:t})$ and
\texttt{complete_conditional()} to compute $p(x_{t+1}\mid
y_{1:t+1})$. As in the normal-gamma example, it is up to the user to
reason about the graphical model structure, but Autoconj handles all
of the conditioning and marginalization automatically. The same code
could be applied to a hidden Markov model (which has the same
graphical model structure) by simply changing the distributions in the
log-joint and the support from real to integer.

When not all complete conditionals are tractable, the variational
evidence lower bound (ELBO) is not tractable to compute exactly.
Several strategies exist for dealing with this problem. One
approach is to find a lower bound on the log-joint that is only a
function of expected sufficient statistics of some exponential family
\citep{jaakkola1996variational,blei2005correlated}. Another is to linearize
problematic terms in the log-joint \citep{khan2015kullback}.

Knowledge of conjugate pairs is not sufficient to implement either of
these strategies, which rely on direct manipulation of the
log-joint to achieve a kind of quasi-conjugacy. But Autoconj naturally
facilitates these strategies, since it does not require that the
log-joint functions it is given exactly correspond to any true generative
process.

\begin{listing}[htb]



\begin{Verbatim}[commandchars=\\\{\},codes={\catcode`\$=3\catcode`\^=7\catcode`\_=8}]
\PYG{k}{def} \PYG{n+nf}{log\PYGZus{}joint\PYGZus{}bound}\PYG{p}{(}\PYG{n}{beta}\PYG{p}{,} \PYG{n}{xi}\PYG{p}{,} \PYG{n}{x}\PYG{p}{,} \PYG{n}{y}\PYG{p}{):}
  \PYG{n}{log\PYGZus{}prior} \PYG{o}{=} \PYG{n}{np}\PYG{o}{.}\PYG{n}{sum}\PYG{p}{(}\PYG{o}{\PYGZhy{}}\PYG{l+m+mf}{0.5} \PYG{o}{*} \PYG{n}{beta}\PYG{o}{**}\PYG{l+m+mi}{2} \PYG{o}{\PYGZhy{}} \PYG{l+m+mf}{0.5} \PYG{o}{*} \PYG{n}{np}\PYG{o}{.}\PYG{n}{log}\PYG{p}{(}\PYG{l+m+mi}{2}\PYG{o}{*}\PYG{n}{np}\PYG{o}{.}\PYG{n}{pi}\PYG{p}{))}
  \PYG{n}{y\PYGZus{}logits} \PYG{o}{=} \PYG{p}{(}\PYG{l+m+mi}{2} \PYG{o}{*} \PYG{n}{y} \PYG{o}{\PYGZhy{}} \PYG{l+m+mi}{1}\PYG{p}{)} \PYG{o}{*} \PYG{n}{np}\PYG{o}{.}\PYG{n}{dot}\PYG{p}{(}\PYG{n}{x}\PYG{p}{,} \PYG{n}{beta}\PYG{p}{)}
  \PYG{c+c1}{\PYGZsh{} Lower bound on \PYGZhy{}log(1 + exp(\PYGZhy{}y\PYGZus{}logits)).}
  \PYG{n}{lamda} \PYG{o}{=} \PYG{p}{(}\PYG{l+m+mf}{0.5} \PYG{o}{\PYGZhy{}} \PYG{n}{expit}\PYG{p}{(}\PYG{n}{xi}\PYG{p}{))} \PYG{o}{/} \PYG{p}{(}\PYG{l+m+mf}{2.} \PYG{o}{*} \PYG{n}{xi}\PYG{p}{)}
  \PYG{n}{log\PYGZus{}likelihood\PYGZus{}bound} \PYG{o}{=} \PYG{n}{np}\PYG{o}{.}\PYG{n}{sum}\PYG{p}{(}\PYG{o}{\PYGZhy{}}\PYG{n}{np}\PYG{o}{.}\PYG{n}{log}\PYG{p}{(}\PYG{l+m+mi}{1} \PYG{o}{+} \PYG{n}{np}\PYG{o}{.}\PYG{n}{exp}\PYG{p}{(}\PYG{o}{\PYGZhy{}}\PYG{n}{xi}\PYG{p}{))} \PYG{o}{+} \PYG{l+m+mf}{0.5} \PYG{o}{*} \PYG{p}{(}\PYG{n}{y\PYGZus{}logits} \PYG{o}{\PYGZhy{}} \PYG{n}{xi}\PYG{p}{)}
                                \PYG{o}{+} \PYG{n}{lamda} \PYG{o}{*} \PYG{p}{(}\PYG{n}{y\PYGZus{}logits} \PYG{o}{**} \PYG{l+m+mi}{2} \PYG{o}{\PYGZhy{}} \PYG{n}{xi} \PYG{o}{**} \PYG{l+m+mi}{2}\PYG{p}{))}
  \PYG{k}{return} \PYG{n}{log\PYGZus{}prior} \PYG{o}{+} \PYG{n}{log\PYGZus{}likelihood\PYGZus{}bound}

\PYG{k}{def} \PYG{n+nf}{xi\PYGZus{}update}\PYG{p}{(}\PYG{n}{beta\PYGZus{}mean}\PYG{p}{,} \PYG{n}{beta\PYGZus{}secondmoment}\PYG{p}{,} \PYG{n}{x}\PYG{p}{):}
  \PYG{l+s+sd}{\PYGZdq{}\PYGZdq{}\PYGZdq{}Sets the bound parameters xi to their optimal value.\PYGZdq{}\PYGZdq{}\PYGZdq{}}
  \PYG{n}{beta\PYGZus{}cov} \PYG{o}{=} \PYG{n}{beta\PYGZus{}secondmoment} \PYG{o}{\PYGZhy{}} \PYG{n}{np}\PYG{o}{.}\PYG{n}{outer}\PYG{p}{(}\PYG{n}{beta\PYGZus{}mean}\PYG{p}{,} \PYG{n}{beta\PYGZus{}mean}\PYG{p}{)}
  \PYG{k}{return} \PYG{n}{np}\PYG{o}{.}\PYG{n}{sqrt}\PYG{p}{(}\PYG{n}{np}\PYG{o}{.}\PYG{n}{einsum}\PYG{p}{(}\PYG{l+s+s1}{\PYGZsq{}ij,ni,nj\PYGZhy{}\PYGZgt{}n\PYGZsq{}}\PYG{p}{,} \PYG{n}{beta\PYGZus{}cov}\PYG{p}{,} \PYG{n}{x}\PYG{p}{,} \PYG{n}{x}\PYG{p}{)} \PYG{o}{+}
                 \PYG{n}{x}\PYG{o}{.}\PYG{n}{dot}\PYG{p}{(}\PYG{n}{beta\PYGZus{}mean}\PYG{p}{)}\PYG{o}{**}\PYG{l+m+mi}{2}\PYG{p}{)}

\PYG{n}{neg\PYGZus{}energy}\PYG{p}{,} \PYG{p}{(}\PYG{n}{t\PYGZus{}beta}\PYG{p}{,),} \PYG{p}{(}\PYG{n}{lognorm\PYGZus{}beta}\PYG{p}{,),} \PYG{o}{=} \PYG{n}{meanfield}\PYG{o}{.}\PYG{n}{multilin\PYGZus{}repr}\PYG{p}{(}
    \PYG{n}{log\PYGZus{}joint\PYGZus{}bound}\PYG{p}{,} \PYG{n}{argnums}\PYG{o}{=}\PYG{p}{(}\PYG{l+m+mi}{0}\PYG{p}{,),} \PYG{n}{supports}\PYG{o}{=}\PYG{p}{(}\PYG{n}{SupportTypes}\PYG{o}{.}\PYG{n}{REAL}\PYG{p}{,),}
    \PYG{n}{example\PYGZus{}args}\PYG{o}{=}\PYG{p}{(}\PYG{n}{beta}\PYG{p}{,} \PYG{n}{xi}\PYG{p}{,} \PYG{n}{x}\PYG{p}{,} \PYG{n}{y}\PYG{p}{))}
\PYG{n}{elbo} \PYG{o}{=} \PYG{n}{partial}\PYG{p}{(}\PYG{n}{meanfield}\PYG{o}{.}\PYG{n}{elbo}\PYG{p}{,} \PYG{n}{neg\PYGZus{}energy}\PYG{p}{,} \PYG{p}{(}\PYG{n}{lognorm\PYGZus{}beta}\PYG{p}{,))}
\PYG{n}{mu\PYGZus{}beta} \PYG{o}{=} \PYG{n}{grad}\PYG{p}{(}\PYG{n}{lognorm\PYGZus{}beta}\PYG{p}{)(}\PYG{n}{grad}\PYG{p}{(}\PYG{n}{neg\PYGZus{}energy}\PYG{p}{)(}\PYG{n}{t\PYGZus{}beta}\PYG{p}{(}\PYG{n}{beta}\PYG{p}{),} \PYG{n}{xi}\PYG{p}{,} \PYG{n}{x}\PYG{p}{,} \PYG{n}{y}\PYG{p}{))}  \PYG{c+c1}{\PYGZsh{} initialize}

\PYG{k}{for} \PYG{n}{iteration} \PYG{o+ow}{in} \PYG{n+nb}{range}\PYG{p}{(}\PYG{l+m+mi}{100}\PYG{p}{):}
  \PYG{n}{xi} \PYG{o}{=} \PYG{n}{xi\PYGZus{}update}\PYG{p}{(}\PYG{n}{mu\PYGZus{}beta}\PYG{p}{[}\PYG{l+m+mi}{0}\PYG{p}{],} \PYG{n}{mu\PYGZus{}beta}\PYG{p}{[}\PYG{l+m+mi}{1}\PYG{p}{],} \PYG{n}{x}\PYG{p}{)}
  \PYG{n}{mu\PYGZus{}beta} \PYG{o}{=} \PYG{n}{grad}\PYG{p}{(}\PYG{n}{lognorm\PYGZus{}beta}\PYG{p}{)(}\PYG{n}{grad}\PYG{p}{(}\PYG{n}{neg\PYGZus{}energy}\PYG{p}{)(}\PYG{n}{mu\PYGZus{}beta}\PYG{p}{,} \PYG{n}{xi}\PYG{p}{,} \PYG{n}{x}\PYG{p}{,} \PYG{n}{y}\PYG{p}{))}
  \PYG{k}{print}\PYG{p}{(}\PYG{l+s+s1}{\PYGZsq{}\PYGZob{}\PYGZcb{}}\PYG{l+s+se}{\PYGZbs{}t}\PYG{l+s+s1}{\PYGZob{}\PYGZcb{}\PYGZsq{}}\PYG{o}{.}\PYG{n}{format}\PYG{p}{(}\PYG{n}{iteration}\PYG{p}{,} \PYG{n}{elbo}\PYG{p}{(}\PYG{n}{mu\PYGZus{}beta}\PYG{p}{,} \PYG{n}{xi}\PYG{p}{,} \PYG{n}{x}\PYG{p}{,} \PYG{n}{y}\PYG{p}{))}
\end{Verbatim}
\caption{
Variational Bayesian logistic regression using the lower bound of
\citet{jaakkola1996variational}. Autoconj can work with
\texttt{log_joint_bound()} even though it is not a true log-joint
density.}
\label{code:logistic-regression}
\end{listing}

Listing~\ref{code:logistic-regression} demonstrates
variational inference for Bayesian logistic regression (which has a non-conjugate
likelihood) using Autoconj to optimize the bound of \citet{jaakkola1996variational}.
One could also use Autoconj to implement other methods such as
proximal variational inference \citep{khan2017cvi,khan2016faster,khan2015kullback}.

\paragraph{Factor Analysis}

\begin{figure}[!tb]
  \centering
  \includegraphics[width=0.49\columnwidth]{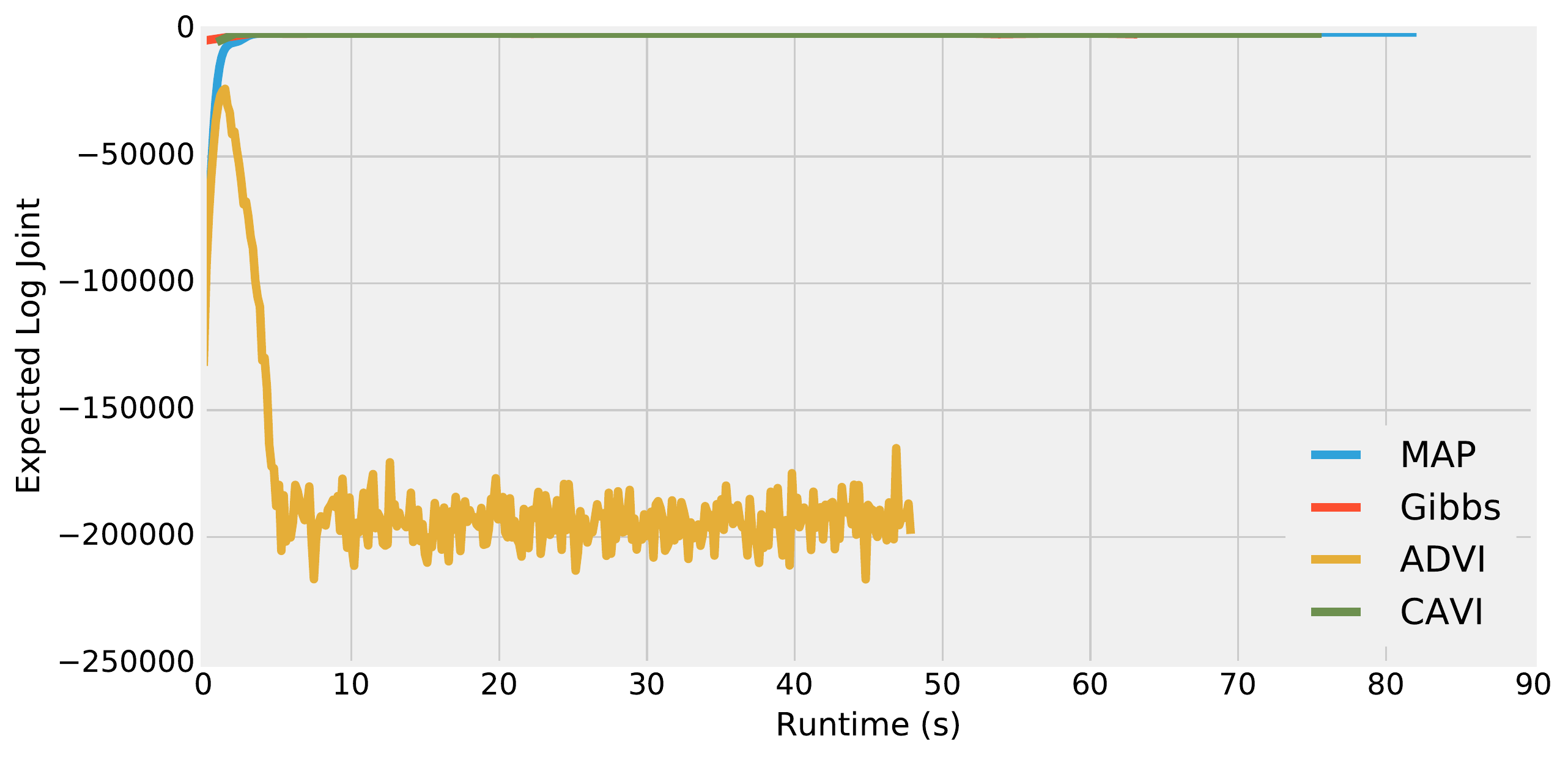}
  \includegraphics[width=0.49\columnwidth]{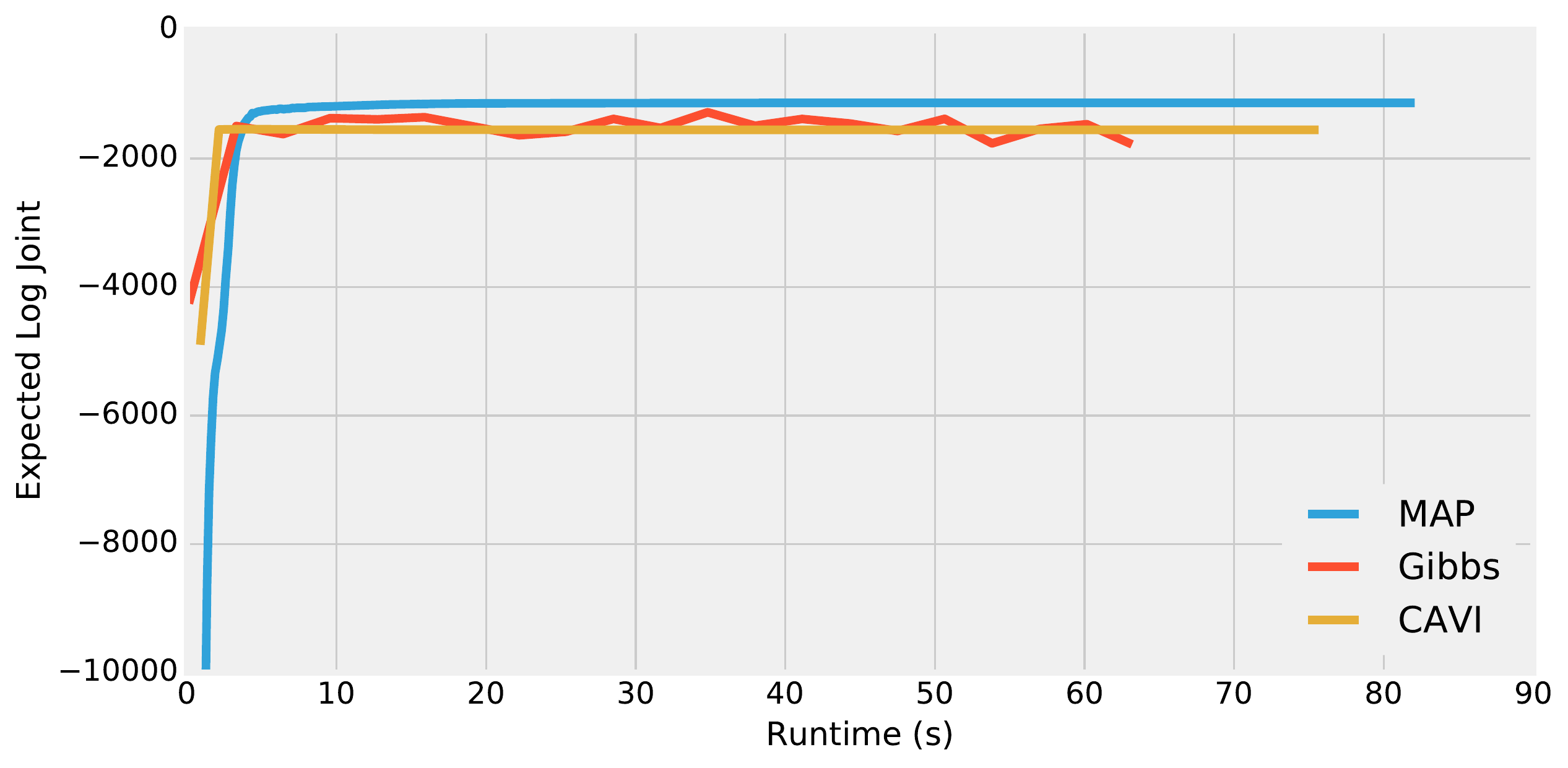}
\caption{
  Comparison of algorithms for Bayesian factor analysis according to their
  estimate of the expected log-joint as a function of runtime.
  \textbf{(left)}
  Relative to other algorithms, mean-field ADVI grossly underfits.
  \textbf{(right)}
  Zoom-in on other algorithms. Block coordinate-ascent variational inference
  (CAVI) converges faster than Gibbs.
}
\label{fig:factor-analysis}
\end{figure}

Autoconj facilitates many structure-exploiting inference algorithms. Here, we
demonstrate why such algorithms are important for efficient inference, and that
Autoconj supports their diverse collection. We generate data from a linear
factor model,
\begin{align*}
w_{mk}\sim \operatorname{Normal}(0, 1);\
z_{nk}\sim \operatorname{Normal}(0, 1);\
\tau\sim \operatorname{Gamma}(\alpha, \beta);\
x_{mn}\sim \operatorname{Normal}(w_m^\top z_n, \tau^{-1/2}).
\end{align*}
There are $N$ examples of $D$-dimensional vectors
$x\in\mathbb{R}^{N\times D}$,
and the data assumes a latent factorization according to
all examples' feature representations
$z\in\mathbb{R}^{N\times K}$
and the principal components
$w\in\mathbb{R}^{D\times K}$.
As a toy demonstration, we use relatively small $N$, $D$, and
$K$.

Autoconj naturally produces a structured mean-field approximation,
since conditioned on $w$ and $x$ the rows of $z$ each have
multivariate-Gaussian complete conditionals (and vice versa for $z$
and $w$). We compared Autoconj structured block coordinate-ascent variational
inference (CAVI) with Autoconj block Gibbs, mean-field ADVI
\citep{kucukelbir2016automatic}, and MAP implemented using scipy.optimize. All
algorithms besides ADVI yield reasonable results, demonstrating the value of
exploiting conjugacy when it is available.

\paragraph{Benchmarking Autoconj}
\label{sub:benchmarking}

\begin{table}[t]
\centering
\begin{tabular}{lr}
\toprule
Implementation & Runtime (s)
\\
\midrule
\textbf{Autoconj (NumPy; 1 CPU)} & \textbf{62.9}  \\
\textbf{Autoconj (TensorFlow; 1 CPU)} & \textbf{75.9}  \\
\textbf{Autoconj (TensorFlow; 6 CPU)} & \textbf{19.7}  \\
\textbf{Autoconj (TensorFlow; 1 GPU)} & \textbf{4.3}  \\
\bottomrule
\end{tabular}
\vspace{1ex}
\caption{Time to run 500 iterations of variational inference on a mixture of Gaussians.
TensorFlow offers little advantage on one CPU core, but an order-of-magnitude speedup on GPU.
}
\label{table:benchmarks}
\end{table}

While we used NumPy as a numerical backend for Autoconj, other
Python-based backends are possible. We wrote a simple translator that
replaces NumPy ops in our computation graph to TensorFlow ops
\citep{abadi2016tensorflow}. We can therefore take a log-joint written
in NumPy, extract complete conditionals or marginals from that model,
and then run the conditional or marginal computations in a TensorFlow
graph (possibly on a GPU or TPU).

We ran Autoconj's CAVI in NumPy and TensorFlow for a
mixture-of-Gaussians model:
\begin{align*}
\pi\sim\operatorname{Dirichlet}(\alpha);\quad
\z_n\sim\operatorname{Categorical}(\pi);\quad
&\mu_{kd}\sim\operatorname{Normal}(0, \sigma);\quad
\tau_{kd}\sim\operatorname{Gamma}(a, b);\quad
\\ &
\x_{nd}\sim\operatorname{Normal}(\mu_{z_n d}, \tau_{z_n d}^{-1/2}).
\end{align*}
See Listing~\ref{code:pplham}.
We automatically translated the NumPy CAVI ops to TensorFlow ops, and
benchmarked 500 iterations of CAVI in NumPy and
TensorFlow on CPU and GPU. Table~\ref{table:benchmarks}
shows the results, which clearly demonstrate the value of running on
GPUs.

\begin{listing}[htb]



\begin{Verbatim}[commandchars=\\\{\},codes={\catcode`\$=3\catcode`\^=7\catcode`\_=8}]
  \PYG{k+kn}{import} \PYG{n+nn}{autoconj.pplham} \PYG{k+kn}{as} \PYG{n+nn}{ph}  \PYG{c+c1}{\PYGZsh{} a simple \PYGZdq{}probabilistic programming language\PYGZdq{}}

  \PYG{k}{def} \PYG{n+nf}{make\PYGZus{}model}\PYG{p}{(}\PYG{n}{alpha}\PYG{p}{,} \PYG{n}{beta}\PYG{p}{):}
    \PYG{k}{def} \PYG{n+nf}{sample\PYGZus{}model}\PYG{p}{():}
      \PYG{l+s+sd}{\PYGZdq{}\PYGZdq{}\PYGZdq{}Generates matrix of shape [num\PYGZus{}examples, num\PYGZus{}features].\PYGZdq{}\PYGZdq{}\PYGZdq{}}
      \PYG{n}{epsilon} \PYG{o}{=} \PYG{n}{ph}\PYG{o}{.}\PYG{n}{norm}\PYG{o}{.}\PYG{n}{rvs}\PYG{p}{(}\PYG{l+m+mi}{0}\PYG{p}{,} \PYG{l+m+mi}{1}\PYG{p}{,} \PYG{n}{size}\PYG{o}{=}\PYG{p}{[}\PYG{n}{num\PYGZus{}examples}\PYG{p}{,} \PYG{n}{num\PYGZus{}latents}\PYG{p}{])}
      \PYG{n}{w} \PYG{o}{=} \PYG{n}{ph}\PYG{o}{.}\PYG{n}{norm}\PYG{o}{.}\PYG{n}{rvs}\PYG{p}{(}\PYG{l+m+mi}{0}\PYG{p}{,} \PYG{l+m+mi}{1}\PYG{p}{,} \PYG{n}{size}\PYG{o}{=}\PYG{p}{[}\PYG{n}{num\PYGZus{}features}\PYG{p}{,} \PYG{n}{num\PYGZus{}latents}\PYG{p}{])}
      \PYG{n}{tau} \PYG{o}{=} \PYG{n}{ph}\PYG{o}{.}\PYG{n}{gamma}\PYG{o}{.}\PYG{n}{rvs}\PYG{p}{(}\PYG{n}{alpha}\PYG{p}{,} \PYG{n}{beta}\PYG{p}{)}
      \PYG{n}{x} \PYG{o}{=} \PYG{n}{ph}\PYG{o}{.}\PYG{n}{norm}\PYG{o}{.}\PYG{n}{rvs}\PYG{p}{(}\PYG{n}{np}\PYG{o}{.}\PYG{n}{dot}\PYG{p}{(}\PYG{n}{epsilon}\PYG{p}{,} \PYG{n}{w}\PYG{o}{.}\PYG{n}{T}\PYG{p}{),} \PYG{l+m+mf}{1.} \PYG{o}{/} \PYG{n}{np}\PYG{o}{.}\PYG{n}{sqrt}\PYG{p}{(}\PYG{n}{tau}\PYG{p}{))}
      \PYG{k}{return} \PYG{p}{[}\PYG{n}{epsilon}\PYG{p}{,} \PYG{n}{w}\PYG{p}{,} \PYG{n}{tau}\PYG{p}{,} \PYG{n}{x}\PYG{p}{]}
    \PYG{k}{return} \PYG{n}{sample\PYGZus{}model}

  \PYG{n}{num\PYGZus{}examples} \PYG{o}{=} \PYG{l+m+mi}{50}
  \PYG{n}{num\PYGZus{}features} \PYG{o}{=} \PYG{l+m+mi}{10}
  \PYG{n}{num\PYGZus{}latents} \PYG{o}{=} \PYG{l+m+mi}{5}
  \PYG{n}{alpha} \PYG{o}{=} \PYG{l+m+mf}{2.}
  \PYG{n}{beta} \PYG{o}{=} \PYG{l+m+mf}{8.}
  \PYG{n}{sampler} \PYG{o}{=} \PYG{n}{make\PYGZus{}model}\PYG{p}{(}\PYG{n}{alpha}\PYG{p}{,} \PYG{n}{beta}\PYG{p}{)}

  \PYG{n}{log\PYGZus{}joint\PYGZus{}fn} \PYG{o}{=} \PYG{n}{ph}\PYG{o}{.}\PYG{n}{make\PYGZus{}log\PYGZus{}joint\PYGZus{}fn}\PYG{p}{(}\PYG{n}{sampler}\PYG{p}{)}
\end{Verbatim}
\caption{Implementing the log joint function for Table~\ref{table:benchmarks}. This
  example also illustrates how Autoconj could be embedded in a probabilistic
  programming language where models are sampling functions and utilities exist
  for tracing their execution (e.g., \citet{tran2018simple}).}
\label{code:pplham}
\end{listing}

\section{Discussion}

  In this paper,
  we proposed a strategy for automatically deriving conjugacy
  relationships. Unlike
  previous systems which focus on relationships between pairs of
  random variables, Autoconj
  operates directly on Python functions that compute log-joint
  distribution functions. This provides support for
  conjugacy-exploiting algorithms in any Python-embedded PPL.  This
  paves the way for accelerating development of novel inference
  algorithms and structure-exploiting modeling strategies.

\textbf{Acknowledgements.}
We thank the anonymous reviewers for their suggestions and Hung Bui
for helpful discussions.

\clearpage
\bibliographystyle{apalike}
\bibliography{main}

\appendix
\input{appendix}

\end{document}

%% file: preamble/preamble.tex
\usepackage[T1]{fontenc}
\usepackage{tgtermes}
\usepackage{amssymb}

\usepackage{amsfonts}
\usepackage{amsmath}
\usepackage{amsthm}
\usepackage{nicefrac}
\usepackage{bm}
\newtheorem{theorem}{Theorem}[section]

\newtheorem{claim}[theorem]{Claim}

\usepackage[english]{babel}
\usepackage[parfill]{parskip}
\usepackage{afterpage}
\usepackage{enumitem}
\usepackage{framed}
\usepackage{xspace}

\usepackage{lineno}

\newcounter{parcount}

\usepackage[usenames,dvipsnames]{xcolor}
\definecolor{shadecolor}{gray}{0.9}

\usepackage{graphicx}
\usepackage[labelfont=bf]{caption}
\usepackage[format=hang]{subcaption}

\usepackage{booktabs, array}

\usepackage[algoruled]{algorithm2e}
\usepackage{listings}
\usepackage{fancyvrb}
\fvset{fontsize=\small}

\usepackage{natbib}
\usepackage[colorlinks,linktoc=all]{hyperref}
\usepackage[all]{hypcap}
\hypersetup{citecolor=MidnightBlue}
\hypersetup{linkcolor=MidnightBlue}
\hypersetup{urlcolor=MidnightBlue}
\usepackage[nameinlink]{cleveref}
\creflabelformat{equation}{#2\textup{#1}#3}  

\usepackage
[acronym,smallcaps,nowarn,section,nogroupskip,nonumberlist]{glossaries}
\glsdisablehyper{}

\lstdefinestyle{mystyle}{
    commentstyle=\color{OliveGreen},
    numberstyle=\tiny\color{black!60},
    stringstyle=\color{BrickRed},
    basicstyle=\ttfamily\scriptsize,
    breakatwhitespace=false,
    breaklines=true,
    captionpos=b,
    keepspaces=true,
    numbers=none,
    numbersep=5pt,
    showspaces=false,
    showstringspaces=false,
    showtabs=false,
    tabsize=2
}
\usepackage{syntax}


%% file: preamble/preamble_acronyms.tex
\newacronym{VI}{vi}{variational inference}
\newacronym{KL}{kl}{Kullback-Leibler}
\newacronym{ELBO}{elbo}{\emph{evidence lower bound}}
\newacronym{MCMC}{mcmc}{Markov chain Monte Carlo}

%% file: preamble/preamble_math.tex
\newcommand{\E}{\mathbb{E}}

\DeclareMathOperator*{\argmax}{arg\,max}

\newcommand{\A}{\mathcal{A}}

\let\L\relax
\newcommand{\L}{\mathcal{L}}
\newcommand{\R}{\mathbb{R}}
\newcommand{\T}{\mathsf{T}}
\newcommand{\given}{\,|\,}   
\newcommand{\param}{;}       

\newcommand{\X}{\mathcal{X}}

\newcommand{\B}{\mathcal{B}}

\newcommand{\x}{\mathrm{x}}
\newcommand{\z}{\mathrm{z}}

\newcommand{\diff}{\mathrm{d}}

\newif\ifbold
\boldfalse  

\ifbold

  \newcommand{\bz}{{\boldsymbol{z}}}

  \newcommand{\bbeta}{{\boldsymbol{\beta}}}

\else

  \newcommand{\bz}{z}

  \newcommand{\bbeta}{{\beta}}

\fi

%% file: preamble/preamble_tikz.tex
\usepackage{tikz}

\pgfdeclarelayer{edgelayer}
\pgfdeclarelayer{nodelayer}
\pgfsetlayers{edgelayer,nodelayer,main}

\definecolor{hexcolor0xbfbfbf}{rgb}{0.749,0.749,0.749}

\tikzset{>=latex}
\tikzstyle{none}   = [inner sep=0pt]
\tikzstyle{line}   = [ -, thick, shorten <=1pt, shorten >=1pt ]
\tikzstyle{arrow}  = [ ->, thick, shorten <=1pt, shorten >=1pt ]
\tikzstyle{ardash} = [ dashed, ->, thick, shorten <=1pt, shorten >=1pt ]

\tikzstyle{empty}=[circle,opacity=0.0,text opacity=1.0,inner sep=0pt]
\tikzstyle{box}=[rectangle,fill=White,draw=Black]
\tikzstyle{filled}=[circle,thick,fill=hexcolor0xbfbfbf,draw=Black]
\tikzstyle{hollow}=[circle,thick,fill=White,draw=Black]
\tikzstyle{param}=[rectangle,fill=Black,draw=Black,inner sep=0pt,minimum width=4pt,minimum height=4pt]
\tikzstyle{paramhollow}=[rectangle,thick,fill=White,draw=Black,inner sep=0pt,minimum
width=4pt,minimum height=4pt]

\usepackage{pgfplots}                               
\pgfplotsset{compat=newest}
\pgfplotsset{plot coordinates/math parser=false}
\newlength\figureheight
\newlength\figurewidth
\newlength\figureheightsmall
\newlength\figurewidthsmall
\definecolor{POSTcolor}{rgb}{0.48, 0.20, 0.58} 
\definecolor{Qcolor}{rgb}{0.00, 0.53, 0.22} 

%% file: appendix.tex
\section{Code Examples}
\label{sec:supp-code}

\subsection{Kalman Filter}
Listing~\ref{code:kalman} demonstrates computing the marginal likelihood
of a time-series $y_{1:T}$ under the linear-Gaussian model
\begin{equation}
\begin{split}
x_1\sim \operatorname{Normal}(0, s_0);\quad
x_{t>1}\sim \operatorname{Normal}(x_{t-1}, s_x);\quad
y_t\sim \operatorname{Normal}(x_{t}, s_y).
\end{split}
\end{equation}

\begin{lstlisting}[language=python,label={code:kalman},
    caption={Exact marginalization in a Kalman filter.}]
def log_p_x1_y1(x1, y1, x1_scale, y1_scale):
  """Computes log p(x_1, y_1)."""
  log_p_x1 = log_probs.norm_gen_log_prob(x1, 0, x1_scale)
  log_p_y1_given_x1 = log_probs.norm_gen_log_prob(y1, x1, y1_scale)
  return log_p_x1 + log_p_y1_given_x1

def log_p_xt_xtt_ytt(xt, xtt, ytt, xt_prior_mean, xt_prior_scale, x_scale,
                     y_scale):
  """Given log p(x_t | y_{1:t}), computes log p(x_t, x_{t+1}, y_{t+1})."""
  log_p_xt = log_probs.norm_gen_log_prob(xt, xt_prior_mean, xt_prior_scale)
  log_p_xtt = log_probs.norm_gen_log_prob(xtt, xt, x_scale)
  log_p_ytt = log_probs.norm_gen_log_prob(ytt, xtt, y_scale)
  return log_p_xt + log_p_xtt + log_p_ytt

def make_marginal_fn():
  # p(x_1 | y_1)
  x1_given_y1_factory = complete_conditional(
      log_p_x1_y1, 0, SupportTypes.REAL, *([1.] * 4))
  # log p(y_1)
  log_p_y1 = marginalize(log_p_x1_y1, 0, SupportTypes.REAL, *([1.] * 4))

  # Given p(x_t | y_{1:t}), compute log p(x_{t+1}, y_{t+1} | y_{1:t}).
  log_p_xtt_ytt = marginalize(
      log_p_xt_xtt_ytt, 0, SupportTypes.REAL, *([1.] * 7))
  # Given p(x_{t+1}, y_{t+1} | y_{1:t}), compute log p(y_{t+1} | y_{1:t}).
  log_p_ytt = marginalize(
      log_p_xtt_ytt, 0, SupportTypes.REAL, *([1.] * 6))
  # Given p(x_{t+1}, y_{t+1} | y_{1:t}), compute p(x_{t+1} | y_{1:t+1}).
  xt_conditional_factory = complete_conditional(
      log_p_xtt_ytt, 0, SupportTypes.REAL, *([1.] * 6))

  def marginal(y_list, x_scale, y_scale):
    # Initialization: compute log p(y_1), p(x_1 | y_1).
    log_p_y = log_p_y1(y_list[0], x_scale, y_scale)
    xt_conditional = x1_given_y1_factory(y_list[0], x_scale, y_scale)

    for t in range(1, len(y_list)):
      # Compute log p(y_t | y_{1:t-1}).
      log_p_y += log_p_ytt(y_list[t], xt_conditional.args[0],
                           xt_conditional.args[1], x_scale, y_scale)
      # Compute p(x_t | y_{1:t}).
      xt_conditional = xt_conditional_factory(
          y_list[t], xt_conditional.args[0], xt_conditional.args[1], x_scale,
          y_scale)
    return log_p_y
  return marginal
\end{lstlisting}